\documentclass[11pt,a4paper]{article}
\usepackage[hyperref]{naacl2019}
\usepackage{times}
\usepackage{latexsym}

\usepackage{longtable}
\usepackage{makecell}

\usepackage{hyperref}
\usepackage{url}
\usepackage{comment}
\usepackage[flushleft]{threeparttable}
\usepackage{graphicx}
\usepackage{amsmath}
\usepackage{subfig}
\usepackage{tabularx}
\usepackage{multirow}
\usepackage{algorithm}
\usepackage{enumitem}
\usepackage{todonotes}

\usepackage{xcolor} 

\newcommand{\Note}[2]{} 
\newcommand{\SideNote}[2]{} 
\renewcommand{\Note}[2]{\todo[color=#1,size=\small, inline=true]{#2}} 
\renewcommand{\SideNote}[2]{\todo[color=#1,size=\small]{#2}} %

\title{On Difficulties of Cross-Lingual Transfer with Order Differences: \\A Case Study on Dependency Parsing}

\aclfinalcopy 

\author{
Wasi Uddin Ahmad$^1$\thanks{~~Equal contribution. Listed by alphabetical order.}, Zhisong Zhang$^2$\footnotemark[1], Xuezhe Ma$^2$\\
\texttt{wasiahmad@cs.ucla.edu,\{zhisongz,xuezhem\}@cs.cmu.edu}\\
\textbf{Eduard Hovy}$^2$, \textbf{Kai-Wei Chang}$^1$, \textbf{Nanyun Peng}$^3$\thanks{~~Corresponding author.}\\
\texttt{ehovy@cs.cmu.edu,kwchang@cs.ucla.edu,npeng@isi.edu} \\ \\
$^1$University of California, Los Angeles, $^2$Carnegie Mellon University \\ $^3$University of Southern California\\
}

\begin{document}

\setlength{\abovedisplayskip}{5pt}
\setlength{\belowdisplayskip}{5pt}
\graphicspath{{images2/}}

\maketitle

\begin{abstract}
Different languages might have different word orders. 
In this paper, we investigate cross-lingual transfer and posit that an order-agnostic model will perform better when transferring to distant foreign languages. 
%
To test our hypothesis, we train dependency parsers on an English corpus and evaluate their transfer performance on 30 other languages. 
Specifically, we compare encoders and decoders based on Recurrent Neural Networks (RNNs) and modified self-attentive architectures. The former relies on sequential information while the latter is more flexible at modeling word order.   
Rigorous experiments and detailed analysis shows that RNN-based architectures transfer well to languages that are close to English, while self-attentive models have better overall cross-lingual transferability and perform especially well on distant languages. 
\end{abstract}

\section{Introduction}


Cross-lingual transfer, which transfers models across languages, has tremendous practical value. It reduces the requirement of annotated data for a target language and is especially useful when the target language is lack of resources. Recently, this technique has been applied to many NLP tasks such as text categorization \cite{zhou2016transfer}, tagging \cite{kim2017cross}, dependency parsing \cite{guo2015cross,guo2016representation} and machine translation \cite{zoph-EtAl:2016:EMNLP2016}.
Despite the preliminary success, transferring across languages is challenging as it requires understanding and handling differences between languages at levels of morphology, syntax, and semantics. It is especially difficult to learn invariant features that can robustly transfer to distant languages. 

Prior work on cross-lingual transfer mainly focused on sharing word-level information by leveraging multi-lingual word embeddings \cite{xiao2014distributed, guo2016representation, sil2017neural}. 
However, words are not independent in sentences; their combinations form larger linguistic units, known as \emph{context}.
Encoding context information is vital for many NLP tasks, and a variety of approaches (e.g., convolutional neural networks and recurrent neural networks) have been proposed to encode context as a high-level feature for downstream tasks. In this paper, we study how to transfer generic contextual information across languages.  
For cross-language transfer, one of the key challenges is the variation in word order among different languages. For example, the Verb-Object pattern in English can hardly be found in Japanese.
This challenge should be taken into consideration in model design.
RNN is a prevalent family of models for many NLP tasks and has demonstrated compelling performances \cite{mikolov2010recurrent,sutskever2014sequence,peters2018deep}. 
However, its sequential nature makes it heavily reliant on word order information, which exposes to the risk of 
encoding language-specific order information that cannot generalize across languages. We characterize this as the ``\emph{order-sensitive}'' property.
%
Another family of models known as ``Transformer'' uses self-attention mechanisms to capture context and was shown to be effective in various NLP tasks \cite{vaswani2017attention,liu2018generating,kitaev2018constituency}. 
With modification in position representations, the self-attention mechanism can be more robust than RNNs to the change of word order.
We refer to this as the ``\emph{order-free}'' property. 

In this work, we posit that \emph{order-free} models have better transferability than \emph{order-sensitive} models because they less suffer from overfitting language-specific word order features.
To test our hypothesis, we first quantify language distance in terms of word order typology, and then systematically study the transferability of order-sensitive and order-free neural architectures on cross-lingual dependency parsing.

We use dependency parsing as a test bed primarily because of the availability of unified annotations across a broad spectrum of languages \cite{ud22}.
Besides, word order typology is found to influence dependency parsing \cite{naseem-barzilay-globerson:2012:ACL2012,tackstrom-mcdonald-nivre:2013:NAACL-HLT,zhang-barzilay:2015:EMNLP,TACL892,aufrant-wisniewski-yvon:2016:COLING}.
Moreover, parsing is a low-level NLP task \cite{hashimoto2016joint} that can benefit many downstream applications \cite{mcclosky2011event, gamallo2012dependency, jie2017efficient}. 

We conduct evaluations on 31 languages across a broad spectrum of language families, as shown in Table~\ref{tab:langs}. 
Our empirical results show that \emph{order-free} encoding and decoding models generally perform better than the \emph{order-sensitive} ones for cross-lingual transfer, especially when the source and target languages are distant.

\section{Quantifying Language Distance} 
\label{sec:wo}

\begin{table}[t]
	\centering
	\small
	\begin{tabular}{>{\centering\arraybackslash}p{1.8cm} | >{\centering\arraybackslash}p{4.75cm}}
		\hline
		Language Families & Languages \\
		\hline
		Afro-Asiatic & Arabic (ar), Hebrew (he)\\
		\hline
		Austronesian & Indonesian (id)\\
		\hline
		IE.Baltic & Latvian (lv)\\
		\hline
		IE.Germanic & Danish (da), Dutch (nl), English (en), German (de), Norwegian (no), Swedish (sv)\\
		\hline
		IE.Indic & Hindi (hi)\\
		\hline
		IE.Latin & Latin (la)\\
		\hline
		IE.Romance & Catalan (ca), French (fr), Italian (it), Portuguese (pt), Romanian (ro), Spanish (es)\\
		\hline
		IE.Slavic & Bulgarian (bg), Croatian (hr), Czech (cs), Polish (pl), Russian (ru), Slovak (sk), Slovenian (sl), Ukrainian (uk)\\
		\hline
		Japanese & Japanese (ja)\\
		\hline
		Korean & Korean (ko)\\
		\hline
		Sino-Tibetan & Chinese (zh)\\
		\hline
		Uralic & Estonian (et), Finnish (fi)\\
		\hline
	\end{tabular}
	\caption{\label{tab:langs} The selected languages grouped by language families. ``IE'' is the abbreviation of Indo-European. }
\end{table}

\begin{figure}[t]
	\centering
	\includegraphics[width=0.5\textwidth]{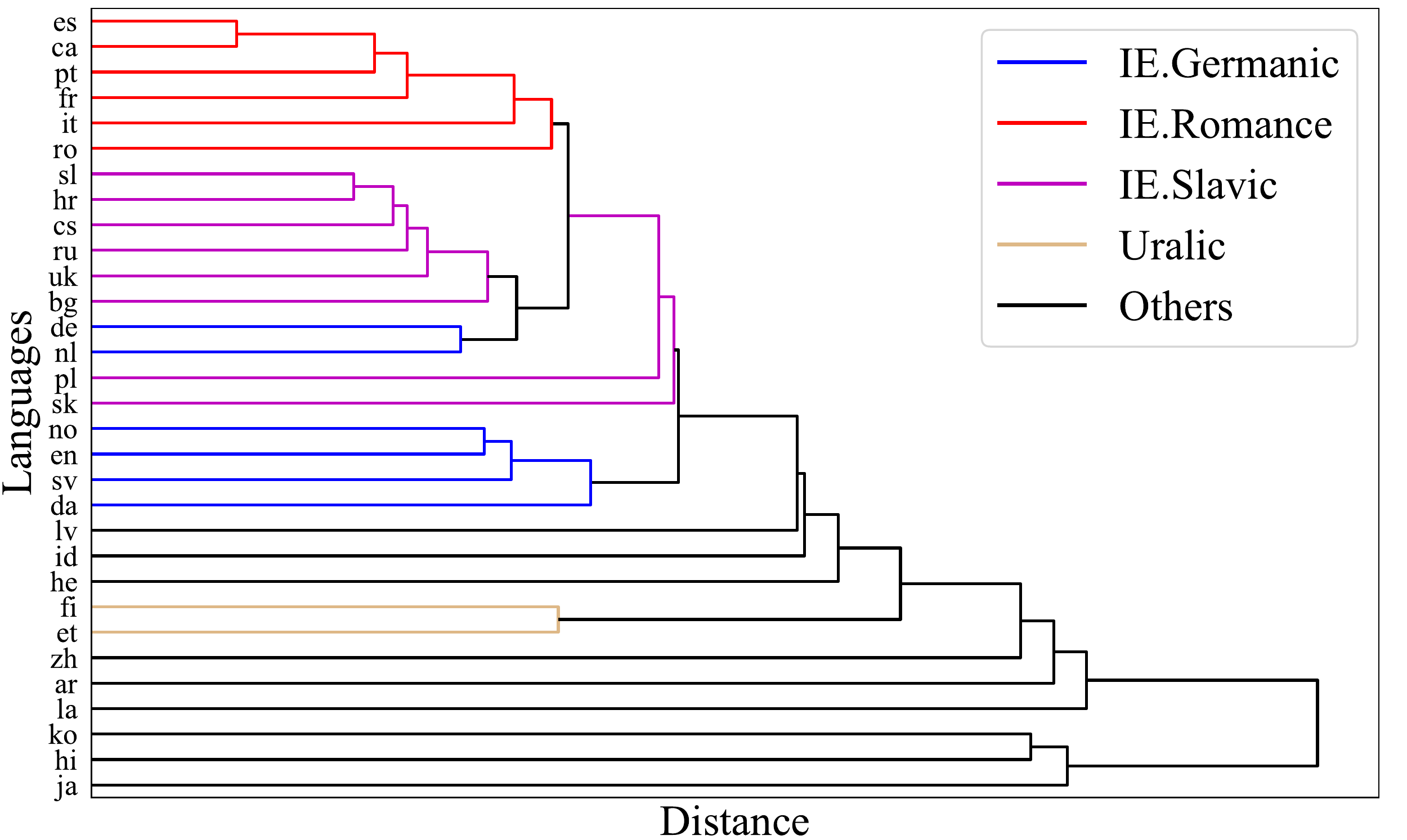}
	\caption{Hierarchical clustering (with the Nearest Point Algorithm) dendrogram of the languages by their word-ordering vectors.}
	\label{fig:lang_cluster}
\end{figure}




We first verify that we can measure ``language distance'' base on word order since it is a significant distinctive feature to differentiate languages \cite{dryer2007word}.
The World Atlas of Language Structures (WALS) \cite{wals} provides a great reference for word order typology and can be used to construct feature vectors for languages \cite{littell-EtAl:2017:EACLshort}. But since we already have the universal dependency annotations, we take an empirical way and directly extract word order features using directed dependency relations \cite{liu2010dependency}.

We conduct our study using the Universal Dependencies (UD) Treebanks (v2.2) \cite{ud22}. 
We select 31 languages for evaluation and analysis, with the selection criterion being that the total token number in the treebanks of that language is over 100K. We group these languages by their language families in Table \ref{tab:langs}. Detailed statistical information of the selected languages and treebanks can be found in Appendix 
A\footnote{Please refer to the supplementary materials for all the appendices of this paper.}.

We look at finer-grained dependency types than the 37 universal dependency labels\footnote{http://universaldependencies.org/u/dep/index.html} in UD v2 
by augmenting the dependency labels with the universal part-of-speech (POS) tags of the head and modifier\footnote{In this paper, we use the term of ``modifier'', which can also be described as ``dependent'' or ``child'' node.} nodes. 
Specifically, we use triples ``(ModifierPOS, HeadPOS, DependencyLabel)'' as the augmented dependency types. 
With this, we can investigate language differences in a fine-grained way by defining directions on these triples (i.e. modifier before head or modifier after head). 

We conduct feature selection by filtering out rare types as they can be unstable.
We defer the results in 52 selected types and more details to Appendix C.
For each dependency type, we collect the statistics of directionality \cite{liu2010dependency,wang2017fine}. Since there can be only two directions for an edge, for each dependency type, we use the relative frequency of the left-direction (modifier before head) as the directional feature. 
By concatenating the directional features of all selected triples, we obtain a word-ordering feature vector for each language. We calculate the \emph{word-ordering distance} using these vectors. In this work, we simply use Manhattan distance, which works well as shown in our analysis (Section \ref{sec:ana}).

We perform hierarchical clustering based on the word-ordering vectors for the selected languages, following \newcite{ostling2015word}. As shown in Figure \ref{fig:lang_cluster}, the grouping of the ground truth language families is almost recovered. The two outliers, German (de) and Dutch (nl), are indeed different from English.
For instance, German and Dutch adopt a larger portion of Object-Verb order in embedded clauses.
The above analysis shows that word order is an important feature to characterize differences between languages. Therefore, it should be taken into consideration in the model design.


\section{Models}
\label{sec:model}






Our primary goal is to conduct cross-lingual transfer of syntactic dependencies without providing any annotation in the target languages. 
The overall architecture of models that are studied in this research is described as follows. The first layer is an input embedding layer, for which we simply concatenate word and POS embeddings. The POS embeddings are trained from scratch, while the word embeddings are fixed and initialized with the multilingual embeddings by \newcite{smith2017offline}.
These inputs are fed to the encoder to get contextual representations, which is further used by the decoder for predicting parse trees.

For the cross-lingual transfer, we hypothesize that the models capturing less language-specific information of the source language will have better transferability. We focus on the word order information, and explore different encoders and decoders that are considered as \emph{order-sensitive} and \emph{order-free}, respectively. 

\subsection{Contextual Encoders}

Considering the sequential nature of languages, RNN is a natural choice for the encoder. However, modeling sentences word by word in the sequence inevitably encodes word order information, which may be specific to the source language. To alleviate this problem, we adopt the self-attention based encoder \cite{vaswani2017attention} for cross-lingual parsing. It can be less sensitive to word order but not necessarily less potent at capturing contextual information, which makes it suitable for our study.

\paragraph{RNNs Encoder}
Following prior work \cite{kiperwasser2016simple,dozat2017biaffiine}, we employ $k$-layer bidirectional LSTMs \cite{hochreiter1997long} on top of the input vectors to obtain contextual representations. Since it explicitly depends on word order, we will refer it as an \emph{order-sensitive} encoder.

\paragraph{Self-Attention Encoder} 
The original self-attention encoder (Transformer) takes absolute positional embeddings as inputs, which capture much order information. To mitigate this, we utilize relative position representations \cite{shaw2018self}, with further simple modification to make it order-agnostic: the original relative position representations discriminate left and right contexts by adding signs to distances, while we discard the directional information. 

We directly base our descriptions on those in \cite{shaw2018self}. For the relative positional self-attention encoder, each layer calculates multiple attention heads. In each head, the input sequence of vectors $\mathbf{x}=(x_1, \dots, x_n)$ are transformed into the output sequence of vectors $\mathbf{z}=(z_1, \dots, z_n)$, based on the self-attention mechanism:
\begin{align*}
z_i &= \sum^{n}_{j=1} \alpha_{ij} (x_j W^V + a_{ij}^V)\\
\alpha_{ij} &= \frac{\exp e_{ij}}{\sum^{n}_{k=1} \exp e_{ik}}\\
e_{ij} &= \frac{x_i W^Q (x_j W^K + a_{ij}^K)^T}{\sqrt{d_z}}
\end{align*}
Here, $a_{ij}^V$ and $a_{ij}^K$ are relative positional representations for the two position $i$ and $j$. Similarly, we clip the distance with a maximum threshold of $k$ (which is empirically set to 10), but we do not discriminate positive and negative values. Instead, since we do not want the model to be aware of directional information, we use the absolute values of the position differences:
\begin{align*}
a_{ij}^K = w_{clip(|j-i|, k)}^K ~~&~~ a_{ij}^V = w_{clip(|j-i|, k)}^V\\
clip(x, k) &= \min(|x|, k)
\end{align*}
Therefore, the learnable relative postion representations have $k+1$ types rather than $2k+1$: we have $w^K=(w_0^K, \dots, w_k^K)$, and $w^V=(w_0^V, \dots, w_k^V)$.

With this, the model knows only what words are surrounding but cannot tell the directions. Since self-attention encoder is less sensitive to word order, we refer to it as an \emph{order-free} encoder.


\subsection{Structured Decoders}
With the contextual representations from the encoder, the decoder predicts the output tree structures. We also investigate two types of decoders with different sensitivity to ordering information.


\paragraph{Stack-Pointer Decoder}
Recently, \citet{ma2018stack} proposed a top-down transition-based decoder and obtained state-of-the-art results. Thus, we select it as our transition-based decoder. To be noted, in this Stack-Pointer decoder, RNN is utilized to record the decoding trajectory and also can be sensitive to word order. Therefore, we will refer to it as an \emph{order-sensitive} decoder.

\paragraph{Graph-based Decoder}
Graph-based decoders assume simple factorization and can search globally for the best structure. Recently, with a deep biaffine attentional scorer, \citet{dozat2017biaffiine} obtained state-of-the-art results with simple first-order factorization \cite{eisner1996three,McDonald:2005}. This method resembles the self-attention encoder and can be regarded as a self-attention output layer. Since it does not depend on ordering information, we refer to it as an \emph{order-free} decoder. 



\section{Experiments and Analysis}
\label{sec:ea}

In this section, we compare four architectures for  cross-lingual transfer dependency parsing with a different combination of order-free and order-sensitive encoder and decoder. We conduct several detailed analyses showing the pros and cons of both types of models.

\begin{table*}[t]
	\centering
	\small
	\begin{tabular}{c@{ }|c@{ }|c@{ }|c@{ }|c@{ }|c@{ }|c@{ }||c}
		\hline
		\multirow{2}{*}{Lang} & {Dist. to} & {SelfAtt-Graph} & {RNN-Graph} & {SelfAtt-Stack} & {RNN-Stack} & Baseline & Supervised \\
		& English & (OF-OF) & (OS-OF) & (OF-OS) & (OS-OS) & \cite{guo2015cross} & (RNN-Graph)\\
		\hline
            en & 0.00 & 
            90.35/88.40 & 90.44/88.31 & 90.18/88.06 & \textbf{91.82}$^\dag$/\textbf{89.89}$^\dag$ & 87.25/85.04 & 90.44/88.31\\
            \hline
            no & 0.06 & 
            80.80/72.81 & 80.67/72.83 & 80.25/72.07 & \textbf{81.75}$^\dag$/\textbf{73.30}$^\dag$ & 74.76/65.16 & 94.52/92.88\\
            sv & 0.07 & 
            80.98/73.17 & 81.23/73.49 & 80.56/72.77 & \textbf{82.57}$^\dag$/\textbf{74.25}$^\dag$ & 71.84/63.52 & 89.79/86.60\\
            fr & 0.09 & 
            77.87/72.78 & \textbf{78.35}$^\dag$/\textbf{73.46}$^\dag$ & 76.79/71.77 & 75.46/70.49 & 73.02/64.67 & 91.90/89.14\\
            pt & 0.09 & 
            \textbf{76.61}$^\dag$/67.75 & 76.46/\textbf{67.98} & 75.39/66.67 & 74.64/66.11 & 70.36/60.11 & 93.14/90.82\\
            da & 0.10 & 
            76.64/67.87 & 77.36/68.81 & 76.39/67.48 & \textbf{78.22}$^\dag$/\textbf{68.83} & 71.34/61.45 & 87.16/84.23\\
            es & 0.12 & 
            74.49/66.44 & \textbf{74.92}$^\dag$/\textbf{66.91}$^\dag$ & 73.15/65.14 & 73.11/64.81 & 68.75/59.59 & 93.17/90.80\\
            it & 0.12 & 
            80.80/75.82 & \textbf{81.10}/\textbf{76.23}$^\dag$ & 79.13/74.16 & 80.35/75.32 & 75.06/67.37 & 94.21/92.38\\
            hr & 0.13 & \textbf{61.91}$^\dag$/\textbf{52.86}$^\dag$ & 60.09/50.67 & 60.58/51.07 & 60.80/51.12 & 52.92/42.19 & 89.66/83.81\\
            ca & 0.13 & 
            73.83/65.13 & \textbf{74.24}$^\dag$/\textbf{65.57}$^\dag$ & 72.39/63.72 & 72.03/63.02 & 68.23/58.15 & 93.98/91.64\\
            pl & 0.13 & \textbf{74.56}$^\dag$/\textbf{62.23}$^\dag$ & 71.89/58.59 & 73.46/60.49 & 72.09/59.75 & 66.74/53.40 & 94.96/90.68\\
            uk & 0.13 & \textbf{60.05}/\textbf{52.28}$^\dag$ & 58.49/51.14 & 57.43/49.66 & 59.67/51.85 & 54.10/45.26 & 85.98/82.21\\
            sl & 0.13 & \textbf{68.21}$^\dag$/\textbf{56.54}$^\dag$ & 66.27/54.57 & 66.55/54.58 & 67.76/55.68 & 60.86/48.06 & 86.79/82.76\\
            nl & 0.14 &
            68.55/60.26 & 67.88/60.11 & 67.88/59.46 & \textbf{69.55}$^\dag$/\textbf{61.55}$^\dag$ & 63.31/53.79 & 90.59/87.52\\
            bg & 0.14 & \textbf{79.40}$^\dag$/\textbf{68.21}$^\dag$ & 78.05/66.68 & 78.16/66.95 & 78.83/67.57 & 73.08/61.23 & 93.74/89.61\\
            ru & 0.14 &
            60.63/51.63 & 59.99/50.81 & 59.36/50.25 & \textbf{60.87}/\textbf{51.96} & 55.03/45.09 & 94.11/92.56\\
            de & 0.14 & \textbf{71.34}$^\dag$/\textbf{61.62}$^\dag$ & 69.49/59.31 & 69.94/60.09 & 69.58/59.64 & 65.14/54.13 & 88.58/83.68\\
            he & 0.14 & \textbf{55.29}/\textbf{48.00}$^\dag$ & 54.55/46.93 & 53.23/45.69 & 54.89/40.95 & 46.03/26.57 & 89.34/84.49\\
            cs & 0.14 & \textbf{63.10}$^\dag$/\textbf{53.80}$^\dag$ & 61.88/52.80 & 61.26/51.86 & 62.26/52.32 & 56.15/44.77 & 94.03/91.87\\
            ro & 0.15 & \textbf{65.05}$^\dag$/\textbf{54.10}$^\dag$ & 63.23/52.11 & 62.54/51.46 & 60.98/49.79 & 56.01/44.04 & 90.07/84.50\\
            sk & 0.17 & \textbf{66.65}/\textbf{58.15}$^\dag$ & 65.41/56.98 & 65.34/56.68 & 66.56/57.48 & 57.75/47.73 & 90.19/86.38\\
            id & 0.17 & \textbf{49.20}$^\dag$/\textbf{43.52}$^\dag$ & 47.05/42.09 & 47.32/41.70 & 46.77/41.28 & 40.84/33.67 & 87.19/82.60\\
            lv & 0.18 & 
            70.78/49.30 & \textbf{71.43}$^\dag$/\textbf{49.59} & 69.04/47.80 & 70.56/48.53 & 62.33/41.42 & 83.67/78.13\\
            fi & 0.20 & 
            66.27/48.69 & \textbf{66.36}/\textbf{48.74} & 64.82/47.50 & 66.25/48.28 & 58.51/38.65 & 88.04/85.04\\
            et & 0.20 & \textbf{65.72}$^\dag$/\textbf{44.87}$^\dag$ & 65.25/44.40 & 64.12/43.26 & 64.30/43.50 & 56.13/34.86 & 86.76/83.28\\
            zh* & 0.23 & \textbf{42.48}$^\dag$/\textbf{25.10}$^\dag$ & 41.53/24.32 & 40.56/23.32 & 40.92/23.45 & 40.03/20.97 & 73.62/67.67\\
            ar & 0.26 & \textbf{38.12}$^\dag$/\textbf{28.04}$^\dag$ & 32.97/25.48 & 32.56/23.70 & 32.85/24.99 & 32.69/22.68 & 86.17/81.83\\
            la & 0.28 & \textbf{47.96}$^\dag$/\textbf{35.21}$^\dag$ & 45.96/33.91 & 45.49/33.19 & 43.85/31.25 & 39.08/26.17 & 81.05/76.33\\
            ko & 0.33 & \textbf{34.48}$^\dag$/\textbf{16.40}$^\dag$ & 33.66/15.40 & 32.75/15.04 & 33.11/14.25 & 31.39/12.70 & 85.05/80.76\\
            hi & 0.40 & \textbf{35.50}$^\dag$/\textbf{26.52}$^\dag$ & 29.32/21.41 & 31.38/23.09 & 25.91/18.07 & 25.74/16.77 & 95.63/92.93\\
            ja* & 0.49 & \textbf{28.18}$^\dag$/\textbf{20.91}$^\dag$ & 18.41/11.99 & 20.72/13.19 & 15.16/9.32 & 15.39/08.41 & 89.06/78.74\\
            \hline
            Average & 0.17 & \textbf{64.06}$^\dag$/\textbf{53.82}$^\dag$ & 62.71/52.63 & 62.22/52.00 & 62.37/51.89 & 57.09/45.41 & 89.44/85.62 \\
		\hline
	\end{tabular}
	\caption{\label{tab:res_test} Results (UAS\%/LAS\%, excluding punctuation) on the test sets. Languages are sorted by the word-ordering distance to English, as shown in the second column. `*' refers to results of delexicalized models, `\dag{}' means that the best transfer model is statistically significantly better (by paired bootstrap test, p $<$ 0.05) than all other transfer models. Models are marked with their encoder and decoder order sensitivity, OF denotes order-free and OS denotes order-sensitive.} 
\vspace{-1em}
\end{table*}


\subsection{Setup}
\paragraph{Settings}
In our main experiments\footnote{Our implementation is publicly available at: https://github.com/uclanlp/CrossLingualDepParser} (those except Section \ref{sec:othersrc}), we take English as the source language and 30 other languages as target languages. We only use the source language for both training and hyper-parameter tuning. 
During testing, we directly apply the trained model to target languages with the inputs from target languages passed through pretrained multilingual embeddings that are projected into a common space as the source language. 
The projection is done by the offline transformation method
\cite{smith2017offline} with pre-trained 300$d$ monolingual embeddings from FastText
\cite{bojanowski2017enriching}. We freeze word embeddings since fine-tuning on them may disturb the multi-lingual alignments. We also adopt gold UPOS tags for the inputs.

For other hyper-parameters, we adopted similar ones as in the Biaffine Graph Parser \cite{dozat2017biaffiine} and the  Stack-Pointer Parser \cite{ma2018stack}. Detailed hyper-parameter settings can be found in Appendix B.
Throughout our experiments, we adopted the language-independent UD labels and a sentence length threshold of 140.
The evaluation metrics are Unlabeled attachment score (UAS) and labeled attachment score (LAS) with punctuations excluded\footnote{In our evaluations, we exclude tokens whose POS tags are ``PUNCT'' or ``SYM''. 
This setting is different from the one adopted in the CoNLL shared task \cite{K18-2001}. However, the patterns are similar as shown in Appendix D where we report the punctuation-included test evaluations.}.
We trained our cross-lingual models five times with different initializations and reported average scores. 

\paragraph{Systems}
As described before, we have an \emph{order-free} (Self-Attention) and an \emph{order-sensitive} (BiLSTM-RNN) encoder, as well as an \emph{order-free} (Biaffine Attention Graph-based) and an \emph{order-sensitive} (Stack-Pointer) decoder. The combination gives us four different models, named in the format of ``Encoder'' plus ``Decoder''. For clarity, we also mark each model with their encoder-decoder order sensitivity characteristics.
For example, ``SelfAtt-Graph (OF-OF)'' refers to the model with self-attention order-free encoder and graph-based order-free decoder. 
We benchmark our models with a baseline shift-reduce transition-based parser, which gave previous state-of-the-art results for single-source zero-resource cross-lingual parsing \cite{guo2015cross}. Since they used older datasets, we re-trained the model on our datasets with their implementation\footnote{https://github.com/jiangfeng1124/acl15-clnndep. We also evaluated our models on the older dataset and compared with their results, as shown in Appendix F.}. We also list the supervised learning results using the ``RNNGraph'' model on each language as a reference of the upper-line for cross-lingual parsing.

\subsection{Results}
\label{sec:results}


The results on the test sets are shown in Table \ref{tab:res_test}. The languages are ordered by their order typology distance to English.
In preliminary experiments, we found our lexicalized models performed poorly on Chinese (zh) and Japanese (ja). We found the main reason was that their embeddings were not well aligned to English. Therefore, we use delexicalized models, where only POS tags are used as inputs. The delexicalized results\footnote{We found delexicalized models to be better only at zh and ja, for about 5 and 10 points respectively. For other languages, they performed worse for about 2 to 5 points. We also tried models without POS, and found them worse for about 10 points on average. We leave further investigation of input representations to future work.
} 
for Chinese and Japanese are listed in the rows marked with ``*''.

Overall, the ``SelfAtt-Graph'' model performs the best in over half of the languages and beats the runner-up ``RNN-Graph'' by around 1.3 in UAS and 1.2 in LAS on average. When compared with ``RNN-Stack'' and ``SelfAtt-Stack'', the average difference is larger than 1.5 points. 
This shows that models capture less word order information generally perform better at cross-lingual parsing. 
Compared with the baseline, our superior results show the importance of the contextual encoder. Compared with the supervised models, the cross-lingual results are still lower by a large gap, indicating space for improvements.

After taking a closer look, we find an interesting pattern in the results: while the model performances on the source language (English) are similar, RNN-based models perform better on languages that are closer to English (upper rows in the table), whereas for languages that are ``distant'' from English, the ``SelfAtt-Graph'' performs much better.
Such patterns correspond well with our hypothesis, that is, the design of models considering word order information is crucial in cross-lingual transfer. 
We conduct more thorough analysis in the next subsection. 

\subsection{Analysis}
\label{sec:ana}

We further analyze how different modeling choices influence cross-lingual transfer. Since we have not touched the training sets for languages other than English, in this subsection, we evaluate and analyze the performance of target languages using training splits in UD. Performance of English is evaluated on the test set. We verify that the trends observed in test set are similar to those on the training sets. As mentioned in the previous section, the bilingual embeddings for Chinese and Japanese do not align well with English. Therefore, we report the results with delexicalizing. 
In the following, we discuss our observations, and detailed results are listed in Appendix E.


\begin{table}[t]
	\centering
	\small
	\begin{tabular}{c|c|c}
		\hline
		Model & UAS\% & LAS\%\\
		\hline
        SelfAtt-Relative (Ours) & 64.57 & 54.14\\
        SelfAtt-Relative+Dir & 63.93 & 53.62\\
        RNN & 63.25 & 52.94\\
        SelfAtt-Absolute & 61.76 & 51.71\\
        SelfAtt-NoPosi & 28.18 & 21.45 \\
		\hline
	\end{tabular}
\caption{\label{tab:posi} Comparisons of different encoders (averaged results over all languages on the original training sets).} 
\vspace{-3mm}
\end{table}
\subsubsection{Encoder Architecture} 
\label{sec:pos_rep_trans}
We assume
models that are less sensitive to word order perform better when transfer to distant languages. To empirically verify this point, we conduct controlled comparisons on various encoders with the same graph-based decoder. Table \ref{tab:posi} shows the average performances in all languages.

To compare models with various degrees of sensitivity to word order, we include several variations of self-attention models. The ``SelfAtt-NoPosi'' is the self-attention model without any positional information. Although it is most insensitive to word order, it performs poorly possibly because of the lack of access to the locality of contexts. The self-attention model with absolute positional embeddings (``SelfAtt-Absolute'') also does not perform well. 
In the case of parsing, relative positional representations may be more useful as indicated by the improvements brought by the directional relative position representations (``SelfAtt-Relative+Dir'') \cite{shaw2018self}. 
Interestingly, the RNN encoder ranks between ``SelfAtt-Relative+Dir'' and ``SelfAtt-Absolute''; all these three encoders explicitly capture word order information in some way. Finally, by discarding the information of directions, our relative position representation (``SelfAtt-Relative'') performs the best (significantly better at p $<$ 0.05). 

One crucial observation we have is that the patterns of breakdown performances for ``SelfAtt-Relative+Dir'' are similar to those of RNN: on closer languages, the direction-aware model performs better, while on distant languages the non-directional one generally obtains better results. Since the only difference between our proposed ``SelfAtt-Relative'' model and the  ``SelfAtt-Relative+Dir'' model is the directional encoding, we believe the better performances should credit to its effectiveness in capturing useful context information without depending too much on the language-specific order information.

These results suggest that a model's sensitivity to word order indeed affects its cross-lingual transfer performances. In later sections, we stick to our ``SelfAtt-Relative'' variation of the self-attentive encoder and focus on the comparisons among the four main models.


\begin{figure}[t]
	\centering
	\hspace{-4mm}
	\includegraphics[width=0.5\textwidth]{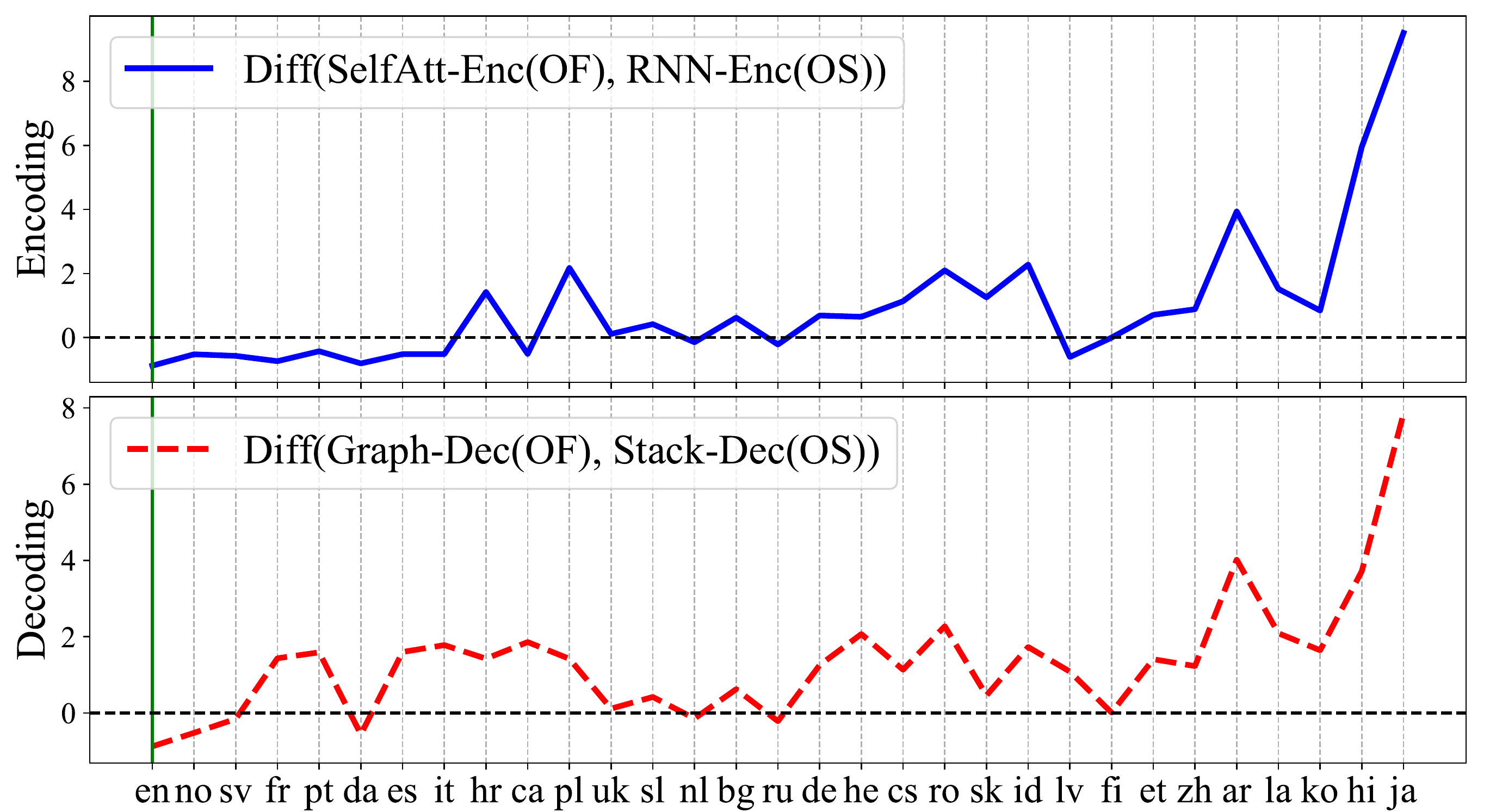}
	\caption{Evaluation score differences between Order-Free (OF) and Order Sensitive (OS) modules. We show results of both encoder (blue solid curve) and decoder (dashed red curve). Languages are sorted by their word-ordering distances to English from left to right. The position of English is marked with a green bar.}
	\label{fig:a2_overall}
\end{figure}

\begin{figure*}[t]
	\centering
	\subfloat[Adposition \& Noun (ADP, NOUN, case)\label{sfig:t_case}]{%
		\includegraphics[width=0.5\textwidth]{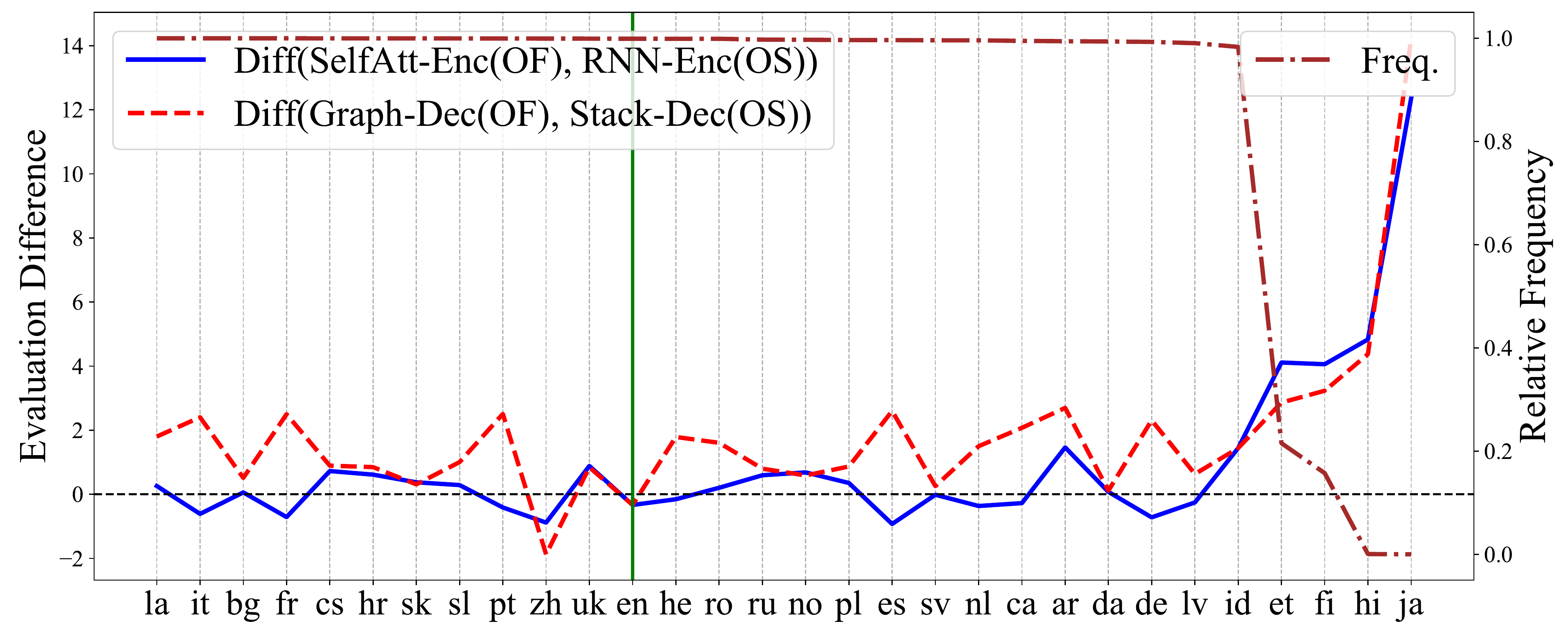}
	}
	\subfloat[Adjective \& Noun (ADJ, NOUN, amod)\label{sfig:t_amod}]{%
		\includegraphics[width=0.5\textwidth]{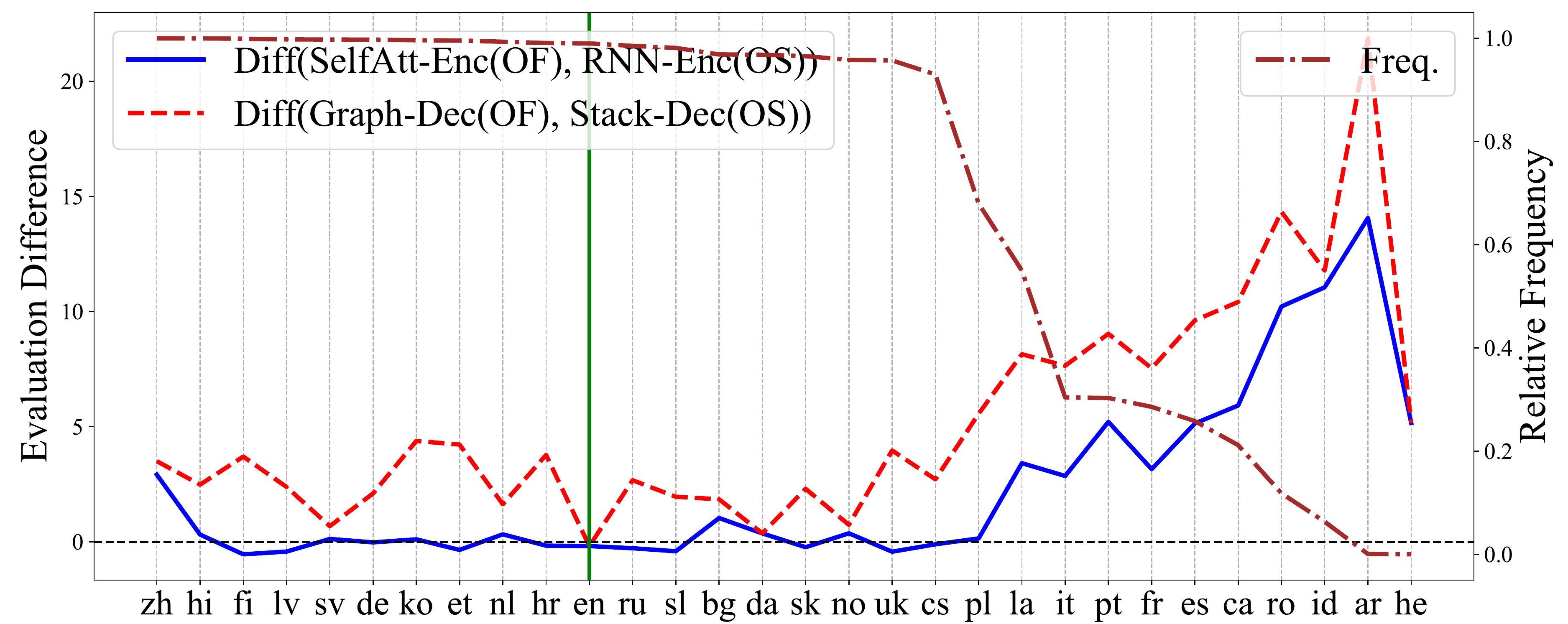}
	}
	
	\subfloat[Auxiliary \& Verb (AUX, VERB, aux)\label{sfig:t_aux}]{%
		\includegraphics[width=0.5\textwidth]{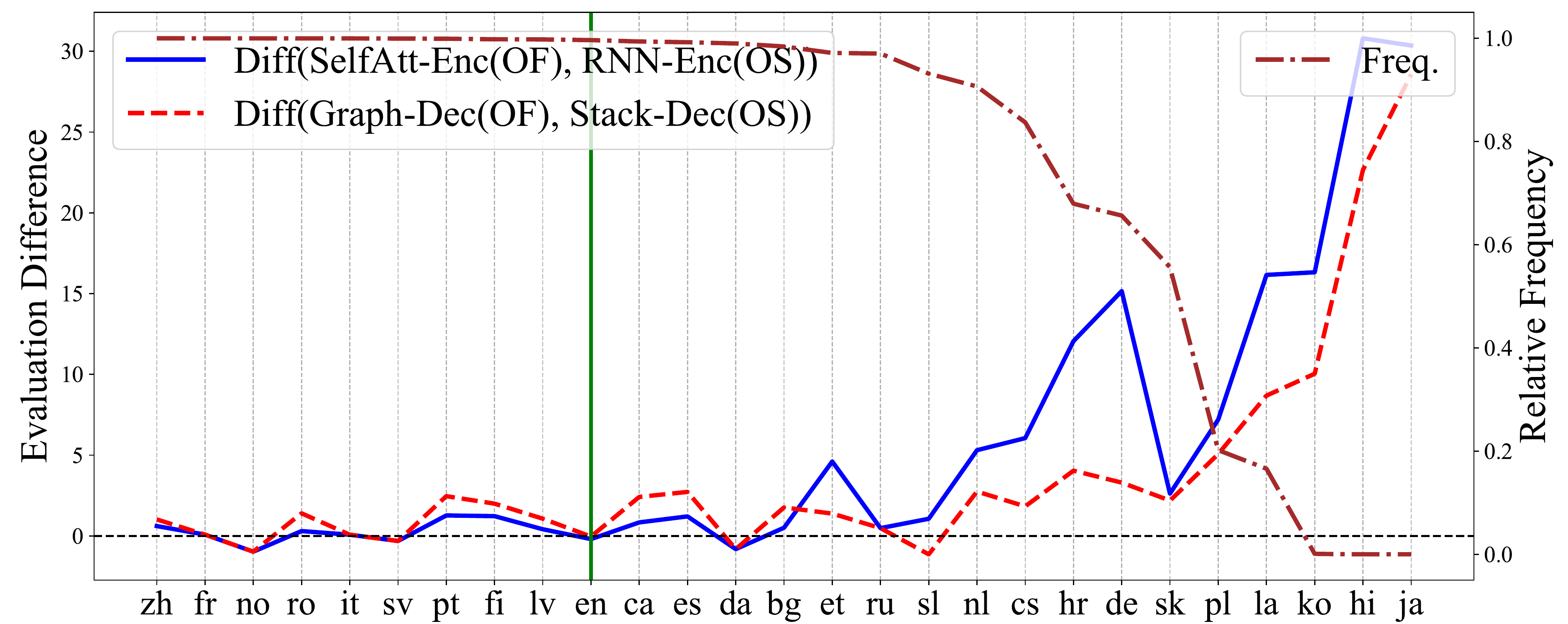}
	}
	\subfloat[Object \& Verb (NOUN, VERB, obj)\label{sfig:t_obj}]{%
		\includegraphics[width=0.5\textwidth]{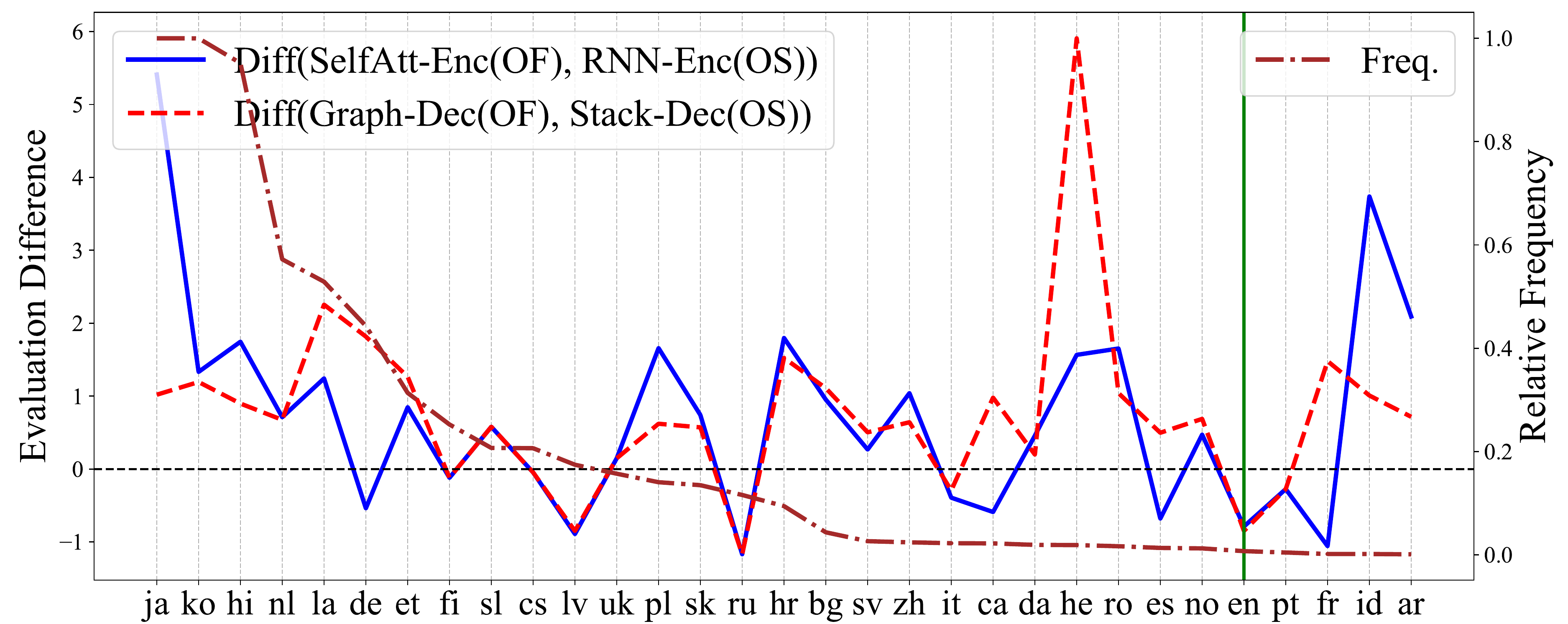}
	}
	\caption{\label{fig:type} Analysis on specific dependency types. To save space, we merge the curves of encoders and decoders into one figure. The blue and red curves and left $y$-axis represent the differences in evaluation scores, the brown curve and right $y$-axis represents the relative frequency of left-direction (modifier before head) on this type. The languages ($x$-axis) are sorted by this relative frequency from high to low. }
	\vspace{-1em}
\end{figure*}

\subsubsection{Performance v.s. Language Distance}
We posit that order-free models can do better than order-sensitive ones on cross-lingual transfer parsing when the target languages have different word orders to the source language. Now we can analyze this with the word-ordering distance.

For each target language, we collect two types of distances when comparing it to English: one is the \emph{word-ordering distance} as described in Section \ref{sec:wo}, the other is the \emph{performance distance}, which is the gap of evaluation scores\footnote{In the rest of this paper, we simply average UAS and LAS for evaluation scores unless otherwise noted.} between the target language and English. The performance distance can represent the general transferability from English to this language. We calculate the correlation of these two distances on all the concerned languages, and the results turn to be quite high: the Pearson and Spearman correlations are \emph{around 0.90 and 0.87} respectively, using the evaluations of any of our four cross-lingual transfer models. This suggests that word order can be an important factor of cross-lingual transferability.

Furthermore, we individually analyze the encoders and decoders of the dependency parsers. Since we have two architectures for each of the modules, when examining one, we take the highest scores obtained by any of the other modules. For example, when comparing RNN and Self-Attention encoders, we take the best evaluation scores of ``RNN-Graph'' and ``RNN-Stack'' for RNN and the best of ``SelfAtt-Graph'' and ``SelfAtt-Stack'' for Self-Attention. Figure \ref{fig:a2_overall} shows the score differences of encoding and decoding architectures against the languages' distances to English.
For both the encoding and decoding module, we observe a similar overall pattern: the order-free models, in general, perform better than order-sensitive ones in the languages that are distant from the source language English. On the other hand, for some languages that are closer to English, order-sensitive models perform better, possibly benefiting from being able to capture similar word ordering information. The performance gap between order-free and order-sensitive models are positively correlated with language distance.

\subsubsection{Performance Breakdown by Types}
Moreover, we compare the results on specific dependency types using concrete examples. For each type, we sort the languages by their relative frequencies of left-direction (modifier before head) and plot the performance differences for encoders and decoders. We highlight the source language English in green. Figure \ref{fig:type} shows four typical example types: Adposition and Noun, Adjective and Noun, Auxiliary and Verb, and Object and Verb. In Figure \ref{sfig:t_case}, we examine the ``case'' dependency type between adpositions and nouns. The pattern is similar to the overall pattern. For languages that mainly use prepositions as in English, different models perform similarly, while for languages that use postpositions, order-free models get better results. The patterns of adjective modifier (Figure \ref{sfig:t_amod}) and auxiliary (Figure \ref{sfig:t_aux}) are also similar.

On dependencies between verbs and object nouns, although in general order-free models perform better, the pattern diverges from what we expect. There can be several possible explanations for this. Firstly, the tokens which are noun objects of verbs only take about 3.1\% on average over all tokens. Considering just this specific dependency type, the correlation between frequency distances and performance differences is 0.64, which is far less than 0.9 when considering all types. Therefore, although Verb-Object ordering is a typical example, we cannot take it as the whole story of word order. Secondly, Verb-Object dependencies can often be difficult to decide. They sometimes are long-ranged and have complex interactions with other words. Therefore, merely reducing modeling order information can have complicated effects.
Moreover, although our relative-position self-attention encoder does not explicitly encode word positions, it may still capture some positional information with relative distances. For example, the words in the middle of a sentence will have different distance patterns from those at the beginning or the end. With this knowledge, the model can still prefer the pattern where a verb is in the middle as in English's Subject-Verb-Object ordering and may find sentences in Subject-Object-Verb languages strange. It will be interesting to explore more ways to weaken or remove this bias.

\subsubsection{Analysis on Dependency Distances}



\begin{table}[t]
	\centering
	\small
	\begin{tabular}{c|c|c}
		\hline
		d & English & Average\\
		\hline
		$<$-2 & 14.36 & 12.93 \\
		-2 & 15.45 & 11.83 \\
		-1 & 31.55 & 30.42 \\
		1 & 7.51 & 14.22 \\
		2 & 9.84 & 10.49 \\
		$>$2 & 21.29 & 20.11 \\
		\hline
	\end{tabular}
\caption{\label{tab:dl} Relative frequencies (\%) of dependency distances. English differs from the Average at $d$=1.}
\vspace{-4mm}
\end{table}

We now look into dependency lengths and directions. Here, we combine dependency length and direction into dependency distance $d$, by using negative signs for dependencies with left-direction (modifier before head) and positive for right-direction (head before modifier). We find a seemingly strange pattern at dependency distances $|d|$=1: for all transfer models, evaluation scores on $d$=-1 can reach about 80, but on $d$=1, the scores are only around 40. This may be explained by the relative frequencies of dependency distances as shown in Table \ref{tab:dl}, where there is a discrepancy between English and the average of other languages at $d$=1. About 80\% of the dependencies with $|d|$=1 in English is the left direction (modifier before head), while overall other languages have more right directions at $|d|$=1. This suggests an interesting future direction of training on more source languages with different dependency distance distributions. 

\begin{figure}[t]
	\centering
	\hspace{-4mm}
	\includegraphics[width=0.5\textwidth]{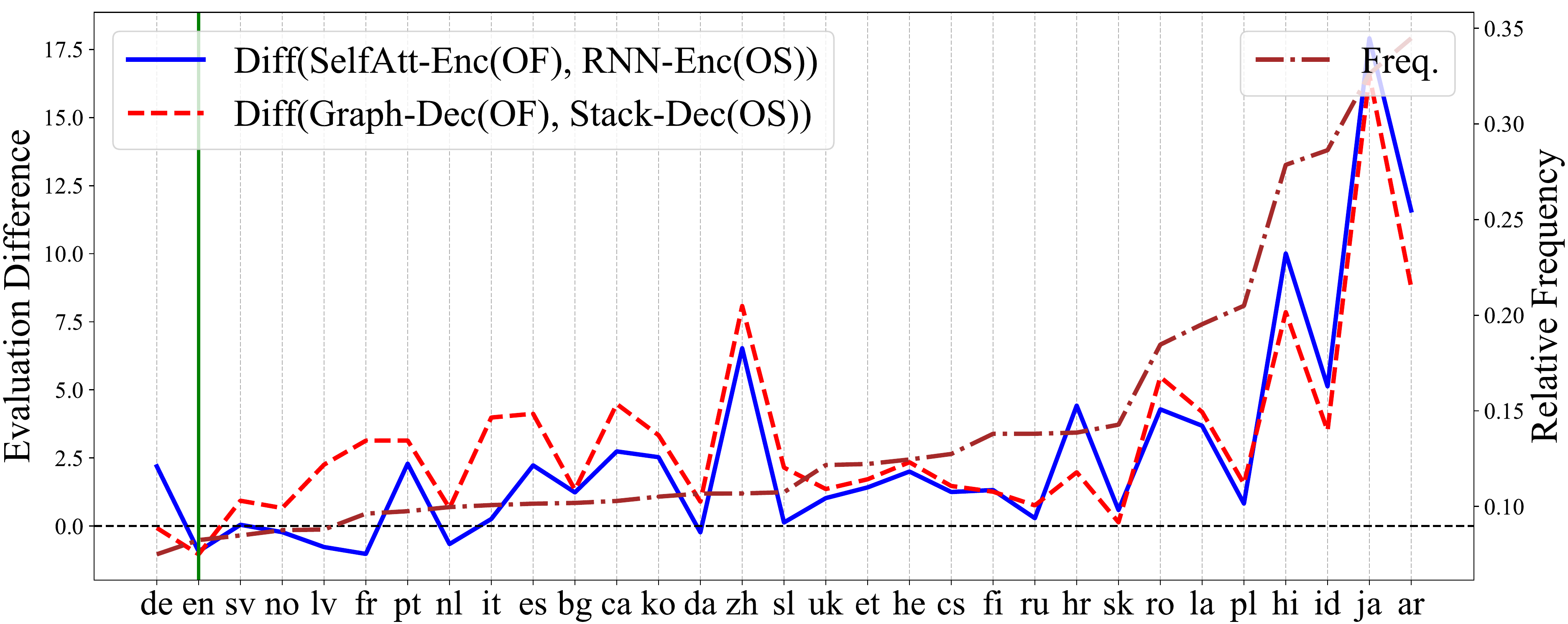}
	\caption{Evaluation differences of models on $d$=1 dependencies. Annotations are the same as in Figure \ref{fig:type}, languages are sorted by percentages (represented by the brown curve and right $y$-axis) of $d$=1 dependencies.}
	\label{fig:a3_cmp}
	\vspace{-3mm}
\end{figure}

We further compare the four models on the $d$=1 dependencies and as shown in Figure \ref{fig:a3_cmp}, the familiar pattern appears again. The order-free models perform better at the languages which have more $d$=1 dependencies. Such finding indicates that our model design of reducing the ability to capture word order information can help on short-ranged dependencies of different directions to the source language.
However, the improvements are still limited.
One of the most challenging parts of unsupervised cross-lingual parsing is modeling cross-lingually shareable and language-unspecific information. In other words, we want flexible yet powerful models. Our exploration of the order-free self-attentive models is the first step.



\begin{figure}[t]
	\centering
	\includegraphics[width=0.5\textwidth]{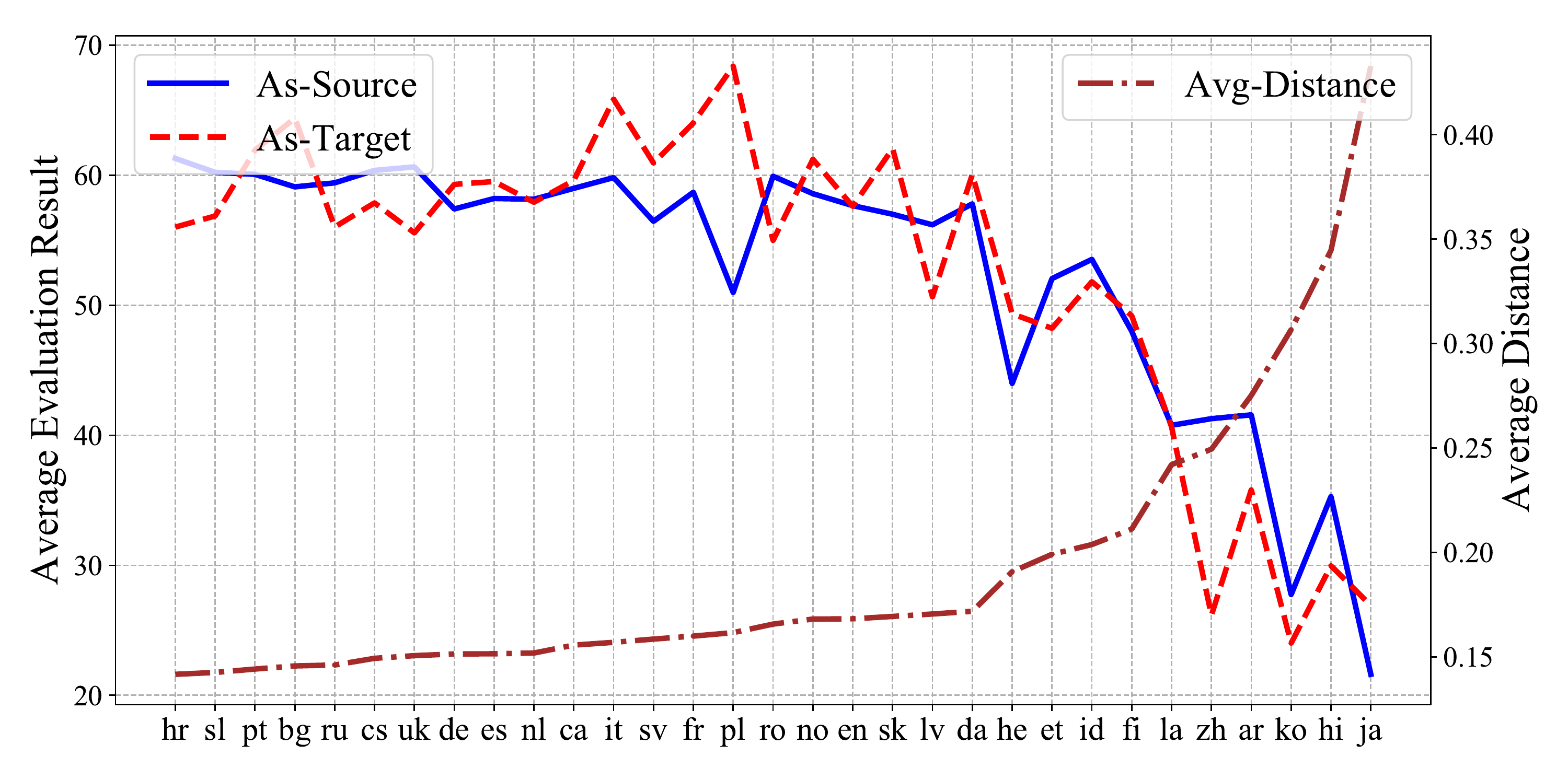}
	\caption{Transfer performance of all source-target language pairs. The blue and red curves show the averages over columns and over rows of the source-target pair performance matrix (see text for details). The brown curve and the right $y$-axis legend represent the average language distance between one language and all others.}
	\label{fig:other}
	\vspace{-5mm}
\end{figure}

\subsubsection{Transfer between All Language Pairs} 
\label{sec:othersrc}

Finally, we investigate the transfer performance of all source-target language pairs.\footnote{Because the size of training corpus for each language is different in UD, to compare among languages, we train a parser on the first 4,000 sentences for each language and evaluate its transfer performance on all other languages.} We first generate a performance matrix $A$, where each entry $(i,j)$ records the transfer performance from a source language $i$ to a target language $j$. We then report the following two aggregate performance measures on $A$ in Figure \ref{fig:other}: 1) \emph{As-source} reports the average over columns of $A$ for each row of the source language and 2) \emph{As-target} reports the average over rows of $A$ for each column of the target language. As a reference, we also plot the average word-order distance between one language to other languages. Results show that both \emph{As-source} (blue line) and \emph{As-target} (red line) highly are anti-correlated (Pearson correlation coefficients are $-0.90$ and $-0.87$, respectively) with average language distance (brown line). 


\section{Related Work}

Cross-language transfer learning employing deep neural networks has widely been studied in the areas of natural language processing \cite{ma-xia:2014:P14-1, guo2015cross, kim2017cross, kann2017one, cotterell2017low}, speech recognition \cite{xu2014cross, huang2013cross}, and information retrieval \cite{vulic2015monolingual, sasaki2018cross, litschko2018unsupervised}.
Learning the language structure (e.g., morphology, syntax)  and transferring knowledge from the source language to the target language is the main underneath challenge, and has been thoroughly investigated for a wide variety of NLP applications, including sequence tagging \cite{yang2016multi, buys2016cross}, name entity recognition \cite{xie2018neural}, 
dependency parsing \cite{tiedemann2015cross, agic2014cross}, entity coreference resolution and linking \cite{kundu2018neural, sil2017neural}, sentiment classification \cite{zhou2015learning, zhou2016cross}, and question answering \cite{joty2017cross}.

Existing work on unsupervised cross-lingual dependency parsing, in general, trains a dependency parser on the source language and then directly run on the target languages.
Training of the monolingual parsers are often delexicalized, i.e., removing all lexical features from the source treebank \cite{zeman2008cross, mcdonald2013universal}, and the underlying feature model is selected from a shared part-of-speech (POS) representation utilizing the Universal POS Tagset \cite{PETROV12.274.L12-1115}.
Another pool of prior work improves the delexicalized approaches by adapting the model to fit the target languages better. 
Cross-lingual approaches that facilitate the usage of lexical features includes choosing the source language data points suitable for the target language \cite{sogaard2011data, tackstrom2013target}, transferring from multiple sources \cite{mcdonald2011multi, guo2016representation, tackstrom2013target}, using cross-lingual word clusters \cite{tackstrom2012cross} and lexicon mapping \cite{xiao2014distributed, guo2015cross}. In this paper, we consider single-source transfer--train a parser on a single source language, and evaluate it on the target languages to test the transferability of neural architectures.

Multilingual transfer \cite{TACL892,naseem-barzilay-globerson:2012:ACL2012,zhang-barzilay:2015:EMNLP} is another broad category of techniques applied to parsing where knowledge from many languages having a common linguistic typology is utilized.
Recent works \cite{aufrant-wisniewski-yvon:2016:COLING, wang2018tacl, wang2018synthetic} demonstrated the significance of explicitly extracting and modeling linguistic properties of the target languages to improve cross-lingual dependency parsing.
Our work is different in that we focus on the neural architectures and explore their influences on cross-lingual transfer.

\section{Conclusion}
In this work, we conduct a comprehensive study on how the design of neural architectures affects cross-lingual transfer learning.
We examine 
two notable families of neural architectures (sequential RNN v.s. self-attention) using dependency parsing as the evaluation task.
We show that \emph{order-free} models perform better than \emph{order-sensitive} ones when there is a significant difference in the word order typology between the target and source language.
In the future, we plan to explore multi-source transfer and incorporating prior linguistic knowledge into the models for better cross-lingual transfer.

\section*{Acknowledgments}
We thank anonymous reviewers for their helpful feedback. We thank Robert {\"O}stling for reaching out when he saw the earlier arxiv version of the paper and providing insightful comments about word order and related citations. We are grateful for the Stanford NLP group's comments and feedback when we present the preliminary results in their seminar. We thank Graham Neubig and the MT/Multilingual Reading Group at CMU-LTI for helpful discussions. We also thank USC Plus Lab and UCLA-NLP group for discussion and comments. This work was supported in part by National Science Foundation Grant IIS-1760523.


\bibliography{naacl2019}
\bibliographystyle{acl_natbib}

\cleardoublepage
\onecolumn

\appendix
\section*{Supplementary Material: Appendices}

\addcontentsline{toc}{section}{Appendices}
\renewcommand{\thesubsection}{\Alph{subsection}}

\subsection{Details of UD Treebanks}
\label{app:langstat}

The statistics of the Universal Dependency treebanks we used are summarized in Table \ref{tab:udstat}.

\begin{small}
	\begin{longtable}{l | l | >{\centering\arraybackslash}p{2cm} | c c c}
		\hline
		Language & Lang. Family & Treebank & & \#Sent. & \#Token(w/o punct)\\
		\hline
		\multirow{3}{*}{Arabic (ar)} & \multirow{3}{*}{Afro-Asiatic} & \multirow{3}{*}{PADT} & train & 6075 & 223881(206041)\\
		& & & dev & 909 & 30239(27339)\\
		& & & test & 680 & 28264(26171)\\
		\hline
		\multirow{3}{*}{Bulgarian (bg)} & \multirow{3}{*}{IE.Slavic} & \multirow{3}{*}{BTB} & train & 8907 & 124336(106813)\\
		& & & dev & 1115 & 16089(13822)\\
		& & & test & 1116 & 15724(13456)\\
		\hline
		\multirow{3}{*}{Catalan (ca)} & \multirow{3}{*}{IE.Romance} & \multirow{3}{*}{AnCora} & train & 13123 & 417587(371981)\\
		& & & dev & 1709 & 56482(50452)\\
		& & & test & 1846 & 57902(51459)\\
		\hline
		\multirow{3}{*}{Chinese (zh)} & \multirow{3}{*}{Sino-Tibetan} & \multirow{3}{*}{GSD} & train & 3997 & 98608(84988)\\
		& & & dev & 500 & 12663(10890)\\
		& & & test & 500 & 12012(10321)\\
		\hline
		\multirow{3}{*}{Croatian (hr)} & \multirow{3}{*}{IE.Slavic} & \multirow{3}{*}{SET} & train & 6983 & 154055(135206)\\
		& & & dev & 849 & 19543(17211)\\
		& & & test & 1057 & 23446(20622)\\
		\hline
		\multirow{3}{*}{Czech (cs)} & \multirow{3}{*}{IE.Slavic} & \multirow{3}{*}{\makecell{PDT,CAC,\\CLTT,FicTree}} & train & 102993 & 1806230(1542805)\\
		& & & dev & 11311 & 191679(163387)\\
		& & & test & 12203 & 205597(174771)\\
		\hline
		\multirow{3}{*}{Danish (da)} & \multirow{3}{*}{IE.Germanic} & \multirow{3}{*}{DDT} & train & 4383 & 80378(69219)\\
		& & & dev & 564 & 10332(8951)\\
		& & & test & 565 & 10023(8573)\\
		\hline
		\multirow{3}{*}{Dutch (nl)} & \multirow{3}{*}{IE.Germanic} & \multirow{3}{*}{\makecell{Alpino,\\LassySmall}} & train & 18058 & 261180(228902)\\
		& & & dev & 1394 & 22938(19645)\\
		& & & test & 1472 & 22622(19734)\\
		\hline
		\multirow{3}{*}{English (en)} & \multirow{3}{*}{IE.Germanic} & \multirow{3}{*}{EWT} & train & 12543 & 204585(180303)\\
		& & & dev & 2002 & 25148(21995)\\
		& & & test & 2077 & 25096(21898)\\
		\hline
		\multirow{3}{*}{Estonian (et)} & \multirow{3}{*}{Uralic} & \multirow{3}{*}{EDT} & train & 20827 & 287859(240496)\\
		& & & dev & 2633 & 37219(30937)\\
		& & & test & 2737 & 41273(34837)\\
		\hline
		\multirow{3}{*}{Finnish (fi)} & \multirow{3}{*}{Uralic} & \multirow{3}{*}{TDT} & train & 12217 & 162621(138324)\\
		& & & dev & 1364 & 18290(15631)\\
		& & & test & 1555 & 21041(17908)\\
		\hline
		\multirow{3}{*}{French (fr)} & \multirow{3}{*}{IE.Romance} & \multirow{3}{*}{GSD} & train & 14554 & 356638(316780)\\
		& & & dev & 1478 & 35768(31896)\\
		& & & test & 416 & 10020(8795)\\
		\hline
		\multirow{3}{*}{German (de)} & \multirow{3}{*}{IE.Germanic} & \multirow{3}{*}{GSD} & train & 13814 & 263804(229338)\\
		& & & dev & 799 & 12486(10809)\\
		& & & test & 977 & 16498(14132)\\
		\hline
		\multirow{3}{*}{Hebrew (he)} & \multirow{3}{*}{Afro-Asiatic} & \multirow{3}{*}{HTB} & train & 5241 & 137680(122122)\\
		& & & dev & 484 & 11408(10050)\\
		& & & test & 491 & 12281(10895)\\
		\hline
		\multirow{3}{*}{Hindi (hi)} & \multirow{3}{*}{IE.Indic} & \multirow{3}{*}{HDTB} & train & 13304 & 281057(262389)\\
		& & & dev & 1659 & 35217(32850)\\
		& & & test & 1684 & 35430(33010)\\
		\hline
		\multirow{3}{*}{Indonesian (id)} & \multirow{3}{*}{Austronesian} & \multirow{3}{*}{GSD} & train & 4477 & 97531(82617)\\
		& & & dev & 559 & 12612(10634)\\
		& & & test & 557 & 11780(10026)\\
		\hline
		\multirow{3}{*}{Italian (it)} & \multirow{3}{*}{IE.Romance} & \multirow{3}{*}{ISDT} & train & 13121 & 276019(244632)\\
		& & & dev & 564 & 11908(10490)\\
		& & & test & 482 & 10417(9237)\\
		\hline
		\multirow{3}{*}{Japanese (ja)} & \multirow{3}{*}{Japanese} & \multirow{3}{*}{GSD} & train & 7164 & 161900(144045)\\
		& & & dev & 511 & 11556(10326)\\
		& & & test & 557 & 12615(11258)\\
		\hline
		\multirow{3}{*}{Korean (ko)} & \multirow{3}{*}{Korean} & \multirow{3}{*}{\makecell{GSD,\\Kaist}} & train & 27410 & 353133(312481)\\
		& & & dev & 3016 & 37236(32770)\\
		& & & test & 3276 & 40043(35286)\\
		\hline
		\multirow{3}{*}{Latin (la)} & \multirow{3}{*}{IE.Latin} & \multirow{3}{*}{PROIEL} & train & 15906 & 171928(171928)\\
		& & & dev & 1234 & 13939(13939)\\
		& & & test & 1260 & 14091(14091)\\
		\hline
		\multirow{3}{*}{Latvian (lv)} & \multirow{3}{*}{IE.Baltic} & \multirow{3}{*}{LVTB} & train & 5424 & 80666(66270)\\
		& & & dev & 1051 & 14585(11487)\\
		& & & test & 1228 & 15073(11846)\\
		\hline
		\multirow{3}{*}{Norwegian (no)} & \multirow{3}{*}{IE.Germanic} & \multirow{3}{*}{\makecell{Bokmaal,\\Nynorsk}} & train & 29870 & 489217(432597)\\
		& & & dev & 4300 & 67619(59784)\\
		& & & test & 3450 & 54739(48588)\\
		\hline
		\multirow{3}{*}{Polish (pl)} & \multirow{3}{*}{IE.Slavic} & \multirow{3}{*}{\makecell{LFG,\\SZ}} & train & 19874 & 167251(136504)\\
		& & & dev & 2772 & 23367(19144)\\
		& & & test & 2827 & 23920(19590)\\
		\hline
		\multirow{3}{*}{Portuguese (pt)} & \multirow{3}{*}{IE.Romance} & \multirow{3}{*}{\makecell{Bosque,\\GSD}} & train & 17993 & 462494(400343)\\
		& & & dev & 1770 & 42980(37244)\\
		& & & test & 1681 & 41697(36100)\\
		\hline
		\multirow{3}{*}{Romanian (ro)} & \multirow{3}{*}{IE.Romance} & \multirow{3}{*}{RRT} & train & 8043 & 185113(161429)\\
		& & & dev & 752 & 17074(14851)\\
		& & & test & 729 & 16324(14241)\\
		\hline
		\multirow{3}{*}{Russian (ru)} & \multirow{3}{*}{IE.Slavic} & \multirow{3}{*}{SynTagRus} & train & 48814 & 870474(711647)\\
		& & & dev & 6584 & 118487(95740)\\
		& & & test & 6491 & 117329(95799)\\
		\hline
		\multirow{3}{*}{Slovak (sk)} & \multirow{3}{*}{IE.Slavic} & \multirow{3}{*}{SNK} & train & 8483 & 80575(65042)\\
		& & & dev & 1060 & 12440(10641)\\
		& & & test & 1061 & 13028(11208)\\
		\hline
		\multirow{3}{*}{Slovenian (sl)} & \multirow{3}{*}{IE.Slavic} & \multirow{3}{*}{\makecell{SSJ,\\SST}} & train & 8556 & 132003(116730)\\
		& & & dev & 734 & 14063(12271)\\
		& & & test & 1898 & 24092(22017)\\
		\hline
		\multirow{3}{*}{Spanish (es)} & \multirow{3}{*}{IE.Romance} & \multirow{3}{*}{\makecell{GSD,\\AnCora}} & train & 28492 & 827053(730062)\\
		& & & dev & 3054 & 89487(78951)\\
		& & & test & 2147 & 64617(56973)\\
		\hline
		\multirow{3}{*}{Swedish (sv)} & \multirow{3}{*}{IE.Germanic} & \multirow{3}{*}{Talbanken} & train & 4303 & 66645(59268)\\
		& & & dev & 504 & 9797(8825)\\
		& & & test & 1219 & 20377(18272)\\
		\hline
		\multirow{3}{*}{Ukrainian (uk)} & \multirow{3}{*}{IE.Slavic} & \multirow{3}{*}{IU} & train & 4513 & 75098(60976)\\
		& & & dev & 577 & 10371(8381)\\
		& & & test & 783 & 14939(12246)\\
		\hline
		\caption{Statistics of the UD Treebanks we used. For language family, ``IE'' is the abbreviation for Indo-European. ``(w/o) punct'' means the numbers of the tokens excluding ``PUNCT'' and ``SYM''.}
		\label{tab:udstat}
	\end{longtable}
\end{small}

\subsection{Hyper-Parameters}
\label{app:hp}

Table \ref{tab:hp} summarizes the hyper-parameters that we used in our experiments. Most of them are similar to those in \cite{dozat2017biaffiine} and \cite{ma2018stack}.

\begin{table}[h]
	\centering
	\small
	\begin{tabular}{c|c|c|c}
		\hline
		 & Layer & Hyper-Parameter & Value\\
		\hline
		\multirow{2}{*}{Input} & Word & dimension & 300\\
		& POS & dimension & 50\\
		\hline
		\multirow{8}{*}{RNN} & \multirow{2}{*}{Encoder} & encoder layer & 3 \\
		& & encoder size & 300\\
		\cline{2-4}
		& \multirow{2}{*}{MLP} & arc MLP size & 512\\
		& & label MLP size & 128\\
		\cline{2-4}
		& \multirow{4}{*}{Training} & Dropout & 0.33\\
		& & optimizer & Adam\\
		& & learning rate & 0.001\\
		& & batch size & 32\\
		\hline
		\multirow{9}{*}{Self-Attention} & \multirow{3}{*}{Encoder} & encoder layer & 6 \\
		& & $d_{model}$ & 350 \\
		& & $d_{ff}$ & 512\\
		\cline{2-4}
		& \multirow{2}{*}{MLP} & arc MLP size & 512\\
		& & label MLP size & 128\\
		\cline{2-4}
		& \multirow{4}{*}{Training} & Dropout & 0.2\\
		& & optimizer & Adam\\
		& & learning rate & 0.0001\\
		& & batch size & 80\\
		\hline
	\end{tabular}
\caption{\label{tab:hp} Hyper-parameters in our experiments. }
\end{table}

\newpage
\subsection{Details about augmented dependency types}
\label{app:dtype}

\begin{table}[h]
	\centering
	\small
	\begin{tabular}{c|c|c|c|c|c}
		\hline
		Type & Avg. Freq. (\%) & \#Lang. & Type & Avg. Freq. (\%) & \#Lang. \\
		\hline
		(ADP, NOUN, case) & 7.47 & 31 & (PROPN, VERB, nsubj) & 0.81 & 30\\
		(PUNCT, VERB, punct) & 6.91 & 30 & (PRON, VERB, obj) & 0.77 & 30\\
		(NOUN, NOUN, nmod) & 4.97 & 31 & (NOUN, ROOT, root) & 0.66 & 31\\
		(ADJ, NOUN, amod) & 4.92 & 31 & (VERB, VERB, xcomp) & 0.61 & 28\\
		(DET, NOUN, det) & 4.69 & 30 & (VERB, VERB, ccomp) & 0.60 & 30\\
		(VERB, ROOT, root) & 4.31 & 31 & (ADP, PRON, case) & 0.57 & 29\\
		(NOUN, VERB, obl) & 3.96 & 30 & (AUX, NOUN, cop) & 0.57 & 28\\
		(NOUN, VERB, obj) & 3.10 & 31 & (ADV, ADJ, advmod) & 0.54 & 29\\
		(NOUN, VERB, nsubj) & 2.89 & 31 & (AUX, ADJ, cop) & 0.50 & 27\\
		(PUNCT, NOUN, punct) & 2.75 & 30 & (PROPN, VERB, obl) & 0.48 & 29\\
		(ADV, VERB, advmod) & 2.43 & 31 & (PRON, VERB, obl) & 0.44 & 30\\
		(AUX, VERB, aux) & 2.29 & 28 & (ADV, NOUN, advmod) & 0.41 & 28\\
		(PRON, VERB, nsubj) & 1.53 & 30 & (ADJ, ROOT, root) & 0.39 & 29\\
		(ADP, PROPN, case) & 1.46 & 29 & (PRON, NOUN, nmod) & 0.39 & 22\\
		(NOUN, NOUN, conj) & 1.32 & 30 & (NOUN, ADJ, obl) & 0.37 & 25\\
		(VERB, NOUN, acl) & 1.31 & 31 & (PROPN, PROPN, conj) & 0.35 & 29\\
		(SCONJ, VERB, mark) & 1.27 & 28 & (NOUN, ADJ, nsubj) & 0.35 & 30\\
		(CCONJ, VERB, cc) & 1.18 & 30 & (CCONJ, ADJ, cc) & 0.29 & 28\\
		(PROPN, NOUN, nmod) & 1.14 & 30 & (PUNCT, NUM, punct) & 0.26 & 24\\
		(CCONJ, NOUN, cc) & 1.13 & 30 & (NOUN, NOUN, nsubj) & 0.25 & 31\\
		(NUM, NOUN, nummod) & 1.11 & 31 & (ADJ, ADJ, conj) & 0.25 & 26\\
		(PROPN, PROPN, flat) & 1.09 & 26 & (CCONJ, PROPN, cc) & 0.22 & 26\\
		(VERB, VERB, conj) & 1.05 & 30 & (PRON, VERB, iobj) & 0.21 & 21\\
		(PUNCT, PROPN, punct) & 0.94 & 29 & (ADV, ADV, advmod) & 0.19 & 21\\
		(VERB, VERB, advcl) & 0.89 & 30 & (NOUN, NOUN, appos) & 0.18 & 23\\
		(PUNCT, ADJ, punct) & 0.89 & 30 & (PROPN, VERB, obj) & 0.17 & 24\\
		\hline
	\end{tabular}
	\caption{\label{tab:dtype} Selected augmented dependency types sorted by their average frequencies. ``\#Lang.'' denotes in how many languages the specific type appears. Since the augmented dependency types can be in hundreds or larger than 1k, but mostly infrequent, we prune them according to average frequency and number of appearing languages. Our pruning criterion is ``$Freq>0.1\%~\text{and}~\#Lang\geq 20$''.}
\end{table}

\newpage
\subsection{Punctuation-included Evaluation on the test sets}
\label{app:ptest}

\begin{table}[h]
	\centering
	\small
	\begin{tabular}{c|c|c|c|c}
		\hline
		Language & SelfAtt-Graph & RNN-Graph &  SelfAtt-Stack & RNN-Stack\\
		\hline
en & 89.29/87.52 & 89.46/87.54 & 89.16/87.26 & \textbf{90.83}/\textbf{89.07} \\
\hline
no & 78.47/71.38 & 78.47/71.50 & 78.11/70.84 & \textbf{79.61}/\textbf{72.10} \\
sv & 79.70/72.69 & 79.94/72.99 & 79.24/72.24 & \textbf{81.44}/\textbf{73.98} \\
fr & 75.58/71.05 & \textbf{76.11}/\textbf{71.79} & 74.32/69.87 & 73.56/69.16 \\
pt & \textbf{73.07}/65.30 & 72.82/\textbf{65.38} & 71.61/63.96 & 71.21/63.76 \\
da & 74.03/66.52 & 74.99/67.67 & 73.76/66.15 & \textbf{75.81}/\textbf{67.76} \\
es & 70.98/63.84 & \textbf{71.50}/\textbf{64.40} & 69.54/62.44 & 69.73/62.37 \\
it & 78.19/73.77 & \textbf{78.63}/\textbf{74.31} & 76.52/72.11 & 78.29/73.84 \\
hr & \textbf{60.58}/\textbf{52.60} & 58.60/50.28 & 59.03/50.65 & 59.27/50.72 \\
ca & 70.47/62.37 & \textbf{70.96}/\textbf{62.85} & 68.91/60.87 & 68.79/60.45 \\
pl & \textbf{74.78}/\textbf{64.68} & 71.73/60.83 & 73.82/63.19 & 72.24/62.11 \\
uk & \textbf{57.57}/\textbf{51.16} & 56.32/50.25 & 54.58/48.18 & 57.31/50.81 \\
sl & \textbf{66.50}/\textbf{55.84} & 64.55/53.84 & 64.83/53.88 & 66.07/55.03 \\
nl & 66.92/59.59 & 66.45/59.54 & 66.05/58.59 & \textbf{68.10}/\textbf{61.01} \\
bg & \textbf{76.15}/\textbf{66.48} & 74.85/65.01 & 74.92/65.23 & 75.69/65.96 \\
ru & 55.85/48.47 & 55.40/47.84 & 54.10/46.62 & \textbf{55.88}/\textbf{48.52} \\
de & \textbf{69.61}/\textbf{61.27} & 67.60/58.86 & 68.18/59.73 & 68.02/59.36 \\
he & \textbf{53.53}/\textbf{46.98} & 53.04/46.16 & 51.53/44.76 & 53.26/40.83 \\
cs & \textbf{60.95}/\textbf{53.03} & 59.56/51.80 & 58.88/50.86 & 59.63/51.13 \\
ro & \textbf{63.11}/\textbf{53.54} & 61.19/51.45 & 60.31/50.63 & 59.38/49.61 \\
sk & \textbf{65.11}/\textbf{57.76} & 63.66/56.38 & 63.68/56.21 & 64.97/57.08 \\
id & \textbf{49.00}/\textbf{44.07} & 47.08/42.78 & 47.03/42.17 & 47.12/42.38 \\
lv & 66.53/49.52 & \textbf{66.95}/\textbf{49.66} & 64.50/47.72 & 65.98/48.46 \\
fi & 64.83/49.83 & \textbf{65.04}/\textbf{49.98} & 63.41/48.61 & 64.97/49.63 \\
et & \textbf{63.50}/\textbf{45.88} & 63.08/45.45 & 61.74/44.12 & 62.15/44.57 \\
zh* & \textbf{40.46}/\textbf{25.52} & 39.54/24.74 & 38.37/23.55 & 39.26/24.25 \\
ar & \textbf{37.15}/\textbf{27.79} & 32.37/25.42 & 31.69/23.46 & 32.04/24.73 \\
la & \textbf{47.96}/\textbf{35.21} & 45.96/33.91 & 45.49/33.19 & 43.85/31.25 \\
ko & \textbf{33.96}/\textbf{17.99} & 33.08/16.96 & 31.68/16.04 & 32.81/16.17 \\
hi & \textbf{36.90}/\textbf{28.52} & 30.94/23.55 & 32.65/24.92 & 26.80/19.49 \\
ja* & \textbf{27.83}/\textbf{21.25} & 18.39/12.59 & 20.33/13.56 & 15.01/9.75 \\
\hline
Average & \textbf{62.21}/\textbf{53.27} & 60.91/52.12 & 60.26/51.34 & 60.62/51.46 \\
		\hline
	\end{tabular}
	\caption{\label{tab:res_train} Evaluations with punctuation included (average UAS\%/LAS\% over 5 runs) on the test sets. The patterns are similar to the punctuation-excluded evaluations in the main content. (Languages are sorted by the word-ordering distance to English, `*' refers to results of delexicalized models.) }
\end{table}

\newpage
\subsection{Results on the original training sets}
\label{app:train}

\begin{table}[h]
	\centering
	\small
	\begin{tabular}{c|c|c|c|c}
		\hline
		Language & SelfAtt-Graph & RNN-Graph &  SelfAtt-Stack & RNN-Stack\\
		\hline
		en$^\circ$ & 90.35/88.40 & 90.44/88.31 & 90.18/88.06 & \textbf{91.82}/\textbf{89.89}\\
		\hline
		no & 80.72/72.45 & 80.59/72.41 & 80.06/71.60 & \textbf{81.46}/\textbf{72.75}\\
		sv & 80.07/71.91 & 80.42/\textbf{72.39} & 79.45/71.28 & \textbf{80.87}/72.25\\
		fr & 79.31/74.73 & \textbf{79.99}/\textbf{75.52} & 78.62/74.02 & 76.84/72.22\\
		pt & 77.06/69.33 & \textbf{77.33}/\textbf{69.91} & 75.84/68.22 & 75.39/67.75\\
		da & 75.75/67.12 & 75.95/67.41 & 75.18/66.55 & \textbf{76.98}/\textbf{67.50}\\
		es & 73.91/66.48 & \textbf{74.39}/\textbf{67.03} & 72.84/65.38 & 72.46/64.78\\
		it & 80.37/75.48 & \textbf{80.89}/\textbf{75.99} & 79.15/74.17 & 79.05/73.91\\
		hr & \textbf{61.57}/\textbf{52.40} & 59.74/50.37 & 59.94/50.43 & 60.44/50.68\\
		ca & 74.40/65.73 & \textbf{74.94}/\textbf{66.21} & 73.01/64.42 & 72.75/63.68\\
		pl & \textbf{75.32}/\textbf{63.26} & 73.12/59.76 & 74.28/61.46 & 73.21/61.02\\
		uk & 65.70/\textbf{57.48} & 64.77/56.40 & 64.10/55.83 & \textbf{65.82}/57.13\\
		sl & \textbf{69.13}/\textbf{58.92} & 67.35/56.87 & 67.74/57.08 & 68.95/58.26\\
		nl & 68.98/60.00 & 68.37/59.52 & 68.22/59.02 & \textbf{69.16}/\textbf{60.11}\\
		bg & \textbf{80.25}/\textbf{68.88} & 78.39/67.03 & 79.19/67.66 & 79.66/68.22\\
		ru & 60.50/51.35 & 59.55/50.17 & 59.01/49.71 & \textbf{60.71}/\textbf{51.57}\\
		de & \textbf{67.23}/\textbf{58.27} & 66.64/57.48 & 66.10/56.89 & 65.88/56.63\\
		he & 58.32/\textbf{49.80} & 57.75/49.07 & 56.36/47.62 & \textbf{58.79}/43.83\\
		cs & \textbf{63.04}/\textbf{53.92} & 61.75/52.91 & 61.11/51.91 & 62.21/52.48\\
		ro & \textbf{65.31}/\textbf{54.22} & 63.17/52.16 & 63.03/51.95 & 61.78/50.52\\
		sk & \textbf{76.07}/\textbf{62.75} & 74.67/61.15 & 75.93/61.97 & 75.37/60.94\\
		id & \textbf{47.92}/\textbf{41.93} & 45.07/39.91 & 46.23/40.16 & 45.62/39.67\\
		lv & 71.69/50.43 & \textbf{72.48}/\textbf{50.85} & 70.24/48.97 & 71.60/49.56\\
		fi & 64.64/46.21 & 64.63/\textbf{46.22} & 63.07/44.82 & \textbf{64.74}/46.09\\
		et & \textbf{66.63}/\textbf{45.58} & 65.78/45.01 & 64.94/44.04 & 65.06/44.33\\
		zh* & \textbf{41.05}/\textbf{23.85} & 40.11/23.02 & 39.49/22.68 & 39.89/22.49\\
		ar & \textbf{38.74}/\textbf{28.24} & 33.66/25.44 & 34.25/24.69 & 33.31/24.86\\
		la & \textbf{49.04}/\textbf{35.48} & 47.12/34.36 & 46.78/33.56 & 45.26/31.97\\
		ko & \textbf{34.62}/\textbf{15.14} & 33.91/14.16 & 32.70/13.77 & 32.95/13.14\\
		hi & \textbf{36.01}/\textbf{27.24} & 29.59/21.75 & 32.02/23.79 & 26.37/18.56\\
		ja* & \textbf{28.19}/\textbf{21.74} & 18.23/12.68 & 20.53/13.78 & 15.21/10.37\\
		%
		\hline
		Average & \textbf{64.57}/\text{54.14} & 63.25/52.94 & 62.88/52.44 & 62.88/52.16\\
		\hline
		\hline
	\end{tabular}
	\caption{\label{tab:res_train} Results (average UAS\%/LAS\% over 5 runs, excluding punctuation) on the original training sets. (Languages are sorted by the word-ordering distance to English, `*' refers to results of delexicalized models, `en$^\circ$' means that for English we use results on the test set since models are trained with the English training set.) }
\end{table}

%
\subsection{Results on Google Universal Dependency Treebanks v2.0}
\label{app:old_data}

We also ran our models on Google Universal Dependency Treebanks v2.0 \cite{mcdonald-EtAl:2013:Short}, which is an older dataset that was used by \cite{guo2015cross}. The results show that our models perform better consistently.

\begin{table}[h]
	\centering
	\small
	\begin{tabular}{c|c|c|c|c|c}
	\hline
	Language & SelfAtt-Graph & RNN-Graph &  SelfAtt-Stack & RNN-Stack & \cite{guo2015cross}\\
	\hline
	German & 65.03/55.03 & 64.60/54.57 & 63.63/54.40 & \textbf{65.51}/\textbf{55.82} & 60.35/51.54\\
	\hline
	French & 74.45/63.28 & \textbf{76.75}/\textbf{65.20} & 73.63/62.76 & 75.13/64.44 & 72.93/63.12\\
	\hline
	Spanish & 72.00/61.50 & 73.99/63.46 & 71.73/61.42 & \textbf{74.13}/\textbf{64.00} & 71.90/62.28\\
	\hline
	\end{tabular}
	\caption{Comparisons (UAS\%/LAS\%) on Google Universal Dependency Treebanks v2.0.}
\end{table}

\newpage
\subsection{Results on specific dependency types for Czech}
\label{app:cs}

In table \ref{tab:cs}, we show results of Czech on some dependency types with evaluation breakdowns on dependency directions. We select Czech mainly for two reasons: (1) It has the largest dataset; (2) Czech is famous for relatively flexible word order. Generally, we can see that models that are more flexible on word ordering perform better. Interestingly, for objective and subjective types, we can see that LAS scores for all models are quite low even when the correct heads are predicted. The reason might be that even the relative-positional self-attention encoder can capture some positional information which further reveals word ordering information in some way.

\begin{table}[h]
	\centering
	\small
	\begin{tabular}{c|c|c|c|c|c}
		\hline
		\multicolumn{6}{c}{(ADP, NOUN, case): (mod-first\% in English is 99.92\%.)}\\
		\hline
		Direction & Percentage & SelfAtt-Graph & RNN-Graph &  SelfAtt-Stack & RNN-Stack\\
		\hline
		mod-first & 99.99\% & \textbf{75.34}/\textbf{75.34} & 74.62/74.61 & 74.46/74.43 & 74.17/74.08\\
		head-first & 0.01\% & -- & -- & -- & --\\
		all & 100.00\% & \textbf{75.33}/\textbf{75.33} & 74.61/74.61 & 74.45/74.43 & 74.17/74.07\\
		\hline
		\hline
		\multicolumn{6}{c}{(NOUN, NOUN, nmod): (mod-first\% in English is 4.72\%.)}\\
		\hline
		Direction & Percentage & SelfAtt-Graph & RNN-Graph &  SelfAtt-Stack & RNN-Stack\\
		\hline
		mod-first & 0.97\% & -- & -- & -- & --\\
		head-first & 99.03\% & 21.38/17.85 & 18.55/16.20 & 20.49/16.61 & \textbf{22.51}/\textbf{19.16}\\
		all & 100.00\% & 21.64/17.68 & 18.86/16.05 & 20.77/16.45 & \textbf{22.78}/\textbf{18.98}\\
		\hline
		\hline
		\multicolumn{6}{c}{(ADJ, NOUN, amod): (mod-first\% in English is 99.01\%.)}\\
		\hline
		Direction & Percentage & SelfAtt-Graph & RNN-Graph &  SelfAtt-Stack & RNN-Stack\\
		\hline
		mod-first & 92.99\% & 88.93/88.92 & \textbf{89.42}/\textbf{89.41} & 85.39/85.21 & 87.26/86.37\\
		head-first & 7.01\% & \textbf{41.80}/\textbf{37.03} & 36.52/32.36 & 34.82/27.19 & 40.59/19.85\\
		all & 100.00\% & 85.63/85.29 & \textbf{85.72}/\textbf{85.41} & 81.85/81.14 & 83.98/81.71\\
		\hline
		\hline
		\multicolumn{6}{c}{(NOUN, VERB, obl): (mod-first\% in English is 9.62\%.)}\\
		\hline
		Direction & Percentage & SelfAtt-Graph & RNN-Graph &  SelfAtt-Stack & RNN-Stack\\
		\hline
		mod-first & 37.80\% & 48.84/40.33 & 46.39/38.49 & 48.75/41.08 & \textbf{50.16}/\textbf{41.64}\\
		head-first & 62.20\% & \textbf{62.81}/\textbf{55.97} & 60.38/53.41 & 62.22/55.37 & 61.73/55.32\\
		all & 100.00\% & \textbf{57.53}/50.06 & 55.09/47.77 & 57.13/49.97 & 57.36/\textbf{50.15}\\
		\hline
		\hline
		\multicolumn{6}{c}{(NOUN, VERB, obj): (mod-first\% in English is 0.72\%.)}\\
		\hline
		Direction & Percentage & SelfAtt-Graph & RNN-Graph &  SelfAtt-Stack & RNN-Stack\\
		\hline
		mod-first & 20.65\% & 55.56/\textbf{0.64} & 53.75/0.46 & 54.08/0.37 & \textbf{60.34}/0.18\\
		head-first & 79.35\% & \textbf{73.18}/\textbf{65.24} & 71.30/62.28 & 72.12/63.81 & 72.76/64.65\\
		all & 100.00\% & 69.54/\textbf{51.90} & 67.68/49.52 & 68.39/50.71 & \textbf{70.20}/51.34\\
		\hline
		\hline
		\multicolumn{6}{c}{(NOUN, VERB, nsubj): (mod-first\% in English is 85.07\%.)}\\
		\hline
		Direction & Percentage & SelfAtt-Graph & RNN-Graph &  SelfAtt-Stack & RNN-Stack\\
		\hline
		mod-first & 60.22\% & \textbf{61.42}/58.33 & 58.12/54.51 & 60.88/58.24 & 60.67/\textbf{58.98}\\
		head-first & 39.78\% & \textbf{64.07}/3.83 & 62.93/3.18 & 62.38/2.97 & 59.94/\textbf{4.42}\\
		all & 100.00\% & \textbf{62.47}/36.65 & 60.03/34.09 & 61.48/36.25 & 60.38/\textbf{37.28}\\
		\hline
		\hline
		\multicolumn{6}{c}{(ADV, VERB, advmod): (mod-first\% in English is 58.82\%.)}\\
		\hline
		Direction & Percentage & SelfAtt-Graph & RNN-Graph &  SelfAtt-Stack & RNN-Stack\\
		\hline
		mod-first & 70.15\% & \textbf{88.23}/\textbf{87.49} & 86.43/85.48 & 86.65/85.30 & 86.64/83.72\\
		head-first & 29.85\% & \textbf{65.79}/\textbf{65.28} & 65.02/64.33 & 65.33/64.35 & 61.93/60.53\\
		all & 100.00\% & \textbf{81.53}/\textbf{80.86} & 80.04/79.17 & 80.29/79.05 & 79.26/76.80\\
		\hline
		\hline
		\multicolumn{6}{c}{(AUX, VERB, aux): (mod-first\% in English is 99.64\%.)}\\
		\hline
		Direction & Percentage & SelfAtt-Graph & RNN-Graph &  SelfAtt-Stack & RNN-Stack\\
		\hline
		mod-first & 83.71\% & 88.78/\textbf{88.19} & 84.44/83.52 & \textbf{89.03}/86.59 & 82.54/76.33\\
		head-first & 16.29\% & \textbf{68.18}/\textbf{65.28} & 54.59/50.87 & 63.96/54.02 & 56.67/20.24\\
		all & 100.00\% & \textbf{85.42}/\textbf{84.46} & 79.57/78.20 & 84.94/81.28 & 78.32/67.19\\
		\hline
		\hline
		\multicolumn{6}{c}{(VERB, VERB, advcl): (mod-first\% in English is 31.02\%.)}\\
		\hline
		Direction & Percentage & SelfAtt-Graph & RNN-Graph &  SelfAtt-Stack & RNN-Stack\\
		\hline
		mod-first & 41.75\% & 57.51/\textbf{55.61} & 56.98/55.60 & \textbf{57.54}/55.03 & 54.74/51.66\\
		head-first & 58.25\% & \textbf{71.52}/\textbf{56.68} & 67.39/56.08 & 67.27/54.17 & 65.93/54.13\\
		all & 100.00\% & \textbf{65.67}/\textbf{56.23} & 63.04/55.88 & 63.21/54.53 & 61.26/53.10\\
		\hline
	\end{tabular}
\caption{\label{tab:cs} Evaluation breakdowns (UAS\%/LAS\%) on dependency directions for Czech on some specific dependency types. ``mod-first'' means the dependency edges whose modifier is before head, ``head-first'' means the opposite, and ``all'' indicates both ``mod-first'' and ``head-first''. ``--'' replaces results that are unstable because of rare appearance (below 1\%).}
\end{table}

\end{document}